\documentclass[sigconf]{acmart}

\usepackage{booktabs} % For formal tables
\usepackage{amsmath}
\usepackage{amssymb}
\usepackage{algorithm}
\usepackage{algorithmic}

\setcopyright{rightsretained}
\acmDOI{10.1145/nnnnnnn.nnnnnnn} % To be updated after completing copyright process

\acmISBN{978-x-xxxx-xxxx-x/YY/MM} % To be updated after completing copyright process

\acmConference[GECCO '19]{the Genetic and Evolutionary Computation Conference 2019}{July 13--17, 2019}{Prague, Czech Republic}
\acmYear{2019}
\copyrightyear{2019}

\acmPrice{15.00}

\begin{document}
\title{Empirical Evaluation of Contextual Policy Search with a Comparison-based
Surrogate Model and Active Covariance Matrix Adaptation}
%\titlenote{Produces the permission block, and copyright information}
%\subtitle{CPS with a Comparison-based Surrogate Model and Active CMA}
%\subtitlenote{The full version of the author's guide is available as \texttt{acmart.pdf} document}

%%% The submitted version for review should be ANONYMOUS
\author{Alexander Fabisch}
%\authornote{}
\orcid{0000-0003-2824-7956}
\affiliation{%
  \institution{Robotics Innovation Center, DFKI GmbH}
  \streetaddress{Robert-Hooke-Straße 1}
  %\city{Bremen} 
  %\state{Bremen} 
  %\postcode{28359}
}
\email{Alexander.Fabisch@dfki.de}

% The default list of authors is too long for headers.
\renewcommand{\shortauthors}{Fabisch}

\begin{abstract}
Contextual policy search (CPS) is a class of multi-task reinforcement learning
algorithms that is particularly useful for robotic applications. A recent
state-of-the-art method is Contextual Covariance Matrix Adaptation Evolution
Strategies (C-CMA-ES). It is based on the standard black-box optimization
algorithm CMA-ES. There are two useful extensions of CMA-ES that we will
transfer to C-CMA-ES and evaluate empirically: ACM-ES, which uses a
comparison-based surrogate model, and aCMA-ES, which uses an active update
of the covariance matrix.
We will show that improvements with these methods can be impressive in terms
of sample-efficiency, although this is not relevant any more for the robotic
domain.
\end{abstract}
%
% The code below should be generated by the tool at
% http://dl.acm.org/ccs.cfm
% Please copy and paste the code instead of the example below. 
%
\maketitle

\section{Introduction}
\label{introduction}

In the domain of robotics, behaviors can be generated with reinforcement
learning \cite{Kober2013}. A standard approach is policy search with
movement primitives. Episodic policy search algorithms are often very similar to
black-box optimization algorithms. Examples are the policy search algorithm
relative entropy policy search \cite{Peters2010} and the black-box optimization
algorithm covariance matrix adaptation evolution strategies (CMA-ES)
\cite{Hansen2001}. The minimal formulation of the problem in policy search is
$$\arg\max_{\boldsymbol{\theta}} \mathbb{E}\left[R(\boldsymbol{\theta})\right],$$
where $\boldsymbol{\theta} \in \mathbb{R}^n$ are typically parameters of a policy
$\pi_{\boldsymbol{\theta}}(\boldsymbol{a}|\boldsymbol{s})$ that have to be optimized and
$R(\boldsymbol{\theta})$ is the return. In general we assume stochastic rewards,
hence, we maximize the expected return. The corresponding formulation of a
black-box optimization problem is
$$\arg \min_{\boldsymbol{x}} f(\boldsymbol{x}),$$
where $f$ is the objective function and $\boldsymbol{x} \in \mathbb{R}^n$ are
parameters of the function. Instead of maximizing the expected return, we
will minimize the objective.

We are interested in an extension to the original policy search formulation
that is called contextual policy search. We seek to optimize
$$\arg \max_{\boldsymbol{\omega}} \int_{\boldsymbol{s}} p(\boldsymbol{s}) \int_{\boldsymbol{\theta}} \pi_{\boldsymbol{\omega}}(\boldsymbol{\theta} | \boldsymbol{s}) \mathbb{E}\left[R(\boldsymbol{\theta}, \boldsymbol{s})\right] d\boldsymbol{\theta} d\boldsymbol{s},$$
where $\boldsymbol{s} \in S$ is a context, $\pi_{\boldsymbol{\omega}}$ is a
stochastic upper-level policy parameterized by $\boldsymbol{\omega}$ that
defines a distribution of policy parameters for a given context
\cite{Deisenroth2013}. The return $R$ is extended to take into account the
context, that is, the context modifies the objective.
During the learning process, we optimize $\boldsymbol{\omega}$, observe the
current context $\boldsymbol{s}$, and select
$\boldsymbol{\theta}_i \sim \pi_{\boldsymbol{\omega}}(\boldsymbol{\theta}|\boldsymbol{s})$.

A corresponding general and deterministic problem formulation is
$$\arg \min_g \int_{\boldsymbol{s}} f_{\boldsymbol{s}}\left(g(\boldsymbol{s})\right) d\boldsymbol{s},$$
where $f_{\boldsymbol{s}}$ is a parameterized objective function and
we want to find an optimal function $g(\boldsymbol{s})$. We call this
the contextual black-box optimization problem.
This, of course, is an extremely difficult problem which is usually relaxed by
restricting the problem to a parameterized class of functions
$g_{\boldsymbol{\omega}}$, often linear functions with a nonlinear projection of
the context, for example, polynomials. The optimization problem becomes
$$\arg\min_{\boldsymbol{\omega}} \int_{\boldsymbol{s}} f_{\boldsymbol{s}}(g_{\boldsymbol{\omega}}(\boldsymbol{s})) d\boldsymbol{s}$$
The challenge of contextual black-box optimization in comparison to black-box
optimization lies in the fact that the true objective function, which is an
integral over all possible context, is not directly available. We can only
sample with a specific context $\boldsymbol{s}$.

Contextual policy search and contextual black-box optimization are
two very similar problem formulations. They correspond to each other like
policy search and black-box optimization. Contextual black-box
optimization can benefit from the ideas of black-box optimization as
contextual policy search can benefit from policy search. We will discuss
connections in the state of the art in the following section.

\section{State of the Art}

Extending standard policy search algorithms to the contextual problem is often
straightforward. For example, reward weighted regression \citep[RWR,][]{Peters2007},
cost-regularized kernel regression \citep[CrKR,][]{Kober2012}, and variational
inference for policy search \citep[VIPS,][]{Neumann2011} directly
support this.

Relative entropy policy search \citep[REPS,][]{Peters2010} has
been used in the contextual setting \cite{Kupcsik2013}. One of the key
advantages of C-REPS over similar methods is that it takes into account that
different contexts might have different reward distributions and computes a
baseline to normalize the reward of each context. Because episodic
REPS can be considered to be a black-box optimization algorithm, this work can
be regarded as a template for the extension of other methods. For example,
Bayesian optimization \cite{Brochu2010} has been extended to Bayesian
optimization for contextual policy search \citep[BO-CPS,][]{Metzen2015},
covariance matrix adaptation evolution strategies \cite{Hansen2001} to its
contextual version C-CMA-ES \cite{Abdolmaleki2017}, and
model-based relative entropy stochastic search \citep[MORE,][]{Abdolmaleki2015}
to C-MORE \cite{Tangkaratt2017}.
There are also variants of these algorithms, for example,
a hybrid of C-REPS and CMA-ES \cite{Abdolmaleki2019}, Contextual REPS has
been extended to support active context selection \cite{Fabisch2014}, and BO-CPS
also has been developed further to actively select contexts in active
contextual entropy search \cite{Metzen2015b} and factored contextual policy
search with Bayesian optimization \cite{Karkus2016}, and the new acquisition
function minimum regret search \cite{Metzen2016} has been developed.
It has been shown that C-REPS, however, is usually not stable and robust against
selection of its hyperparameters \cite{Fabisch2015} and suffers from premature
convergence \cite{Abdolmaleki2017b}.

C-CMA-ES, C-MORE, and BO-CPS can be considered state of the art in contextual
policy search or contextual black-box optimization. BO-CPS is usually only
computationally efficient enough for a small number of parameters, C-CMA-ES is
better if more parameters have to be optimized, and C-MORE can be considered
state of the art for high-dimensional, redundant context vectors because it uses
dimensionality reduction.

In this work, we will build on one of the most promising algorithm: C-CMA-ES.
It is sample-efficient, more computationally efficient
than BO-CPS, has only a few critical hyperparameters with good default values,
and does not suffer from premature convergence like C-REPS.
C-CMA-ES is based on CMA-ES. CMA-ES is an established black-box optimization
algorithm for which many extensions have been developed. We will investigate
two of them in a contextual black-box optimization setting.

\section{Methods}

We describe C-CMA-ES and the two extensions active C-CMA-ES
and C-ACM-ES, which uses a surrogate model.

\subsection{C-CMA-ES}

\begin{algorithm}[tb!]
\caption{Contextual CMA-ES}
\label{alg:CCMAES}
\begin{algorithmic}[1]
\REQUIRE update frequency $\lambda$
and number of samples used for the update $\mu$,
context transformations $\phi(\boldsymbol{s}), \psi(\boldsymbol{s})$,
regularization coefficient $\gamma$,
parameter dimension $n$ and context dimension $n_s$,
initial search distribution defined by $\boldsymbol{W}^0=\boldsymbol{\Sigma}^0=\boldsymbol{I}$ and $\sigma^0$
\STATE $t \gets 1$
\WHILE{not converged}
\FORALL{$i \in \{1, \ldots, \lambda\}$}
\STATE Observe $\boldsymbol{s}_i$
\STATE $\boldsymbol{\theta}_i \sim \mathcal{N}({\boldsymbol{W}^t}^T\phi(\boldsymbol{s}_i), (\sigma^t)^2 {\bf \Sigma}^t)$
\STATE Obtain $R(\boldsymbol{s}_i, \boldsymbol{\theta}_i)$
  %\COMMENT{Return of $\pi_{{\bf \theta}_i}$ in context $\boldsymbol{s}_i$}
\ENDFOR

\STATE Build ${\bf \Phi}$, ${\bf \Psi}$, ${\bf \Theta}$, and ${\bf R}$, where ${\bf \Phi}_i = \phi(\boldsymbol{s}_i)^T$,  ${\bf \Psi}_i = \psi(\boldsymbol{s}_i)^T$,${\bf \Theta}_i = \boldsymbol{\theta}_i^T$, ${\bf R}_i = R(\boldsymbol{s}_i, \boldsymbol{\theta}_i)$
\STATE ${\bf B}^{t} \gets \left( {\bf \Psi^T \Psi} + \gamma {\bf I}\right)^{-1} {\bf \Psi^T R}$
  \COMMENT{Baseline}
\FORALL{$i \in \{1, \ldots, \lambda\}$}
\STATE $\hat{A}(\boldsymbol{s}_i, \boldsymbol{\theta}_i) \gets R(\boldsymbol{s}_i, \boldsymbol{\theta}_i) - {\boldsymbol{B}^{t}}^T\psi(\boldsymbol{s}_i)$
\ENDFOR

\STATE Order $\left[ \left( \boldsymbol{s}_1, \boldsymbol{\theta}_1, \hat{A}(\boldsymbol{s}_1, \boldsymbol{\theta}_1) \right), ... \right]$ descending by advantage values $\hat{A}(\boldsymbol{s}_i, \boldsymbol{\theta}_i)$
\STATE $\boldsymbol{D}_{ij} \gets \frac{\delta_{ij}}{Z} \max(0, \left(\log \mu + 0.5 \right) - \log \left( i \right)$)\\
  \COMMENT{$Z$ is chosen so that weights sum up to one, $\boldsymbol{D}$ is diagonal}
\STATE ${\bf W}^{t+1} \gets \left( {\bf \Phi^T D \Phi} + \gamma {\bf I}\right)^{-1} {\bf \Phi^T D \Theta}$
\STATE $\hat{\boldsymbol{\phi}} = \frac{1}{\lambda} \sum_{i=1}^\lambda \phi(\boldsymbol{s}_i)$; ${\bf y} = \frac{{\bf W}^{t+1} \hat{\boldsymbol{\phi}} - {\bf W}^t \hat{\boldsymbol{\phi}}}{\sigma^t}$
  %\COMMENT{$\sigma$-normalized step}

\STATE ${\bf p}_\sigma^{t+1} \gets \left( 1 - c_{\sigma} \right) {\bf p}_\sigma^t + \sqrt{c_\sigma (2 - c_\sigma) \mu_{eff}} {\left({\bf \Sigma}^t\right)}^{-\frac{1}{2}} {\bf y}$
  %\COMMENT{Step size evolution path}
\STATE $h_\sigma \gets \begin{cases} 1 & \textrm{if }
    \frac{||{\bf p}_\sigma^{t+1}||^2}
         {n \sqrt{1 - (1 - c_\sigma)^{2 t}}}
    < 2 + \frac{4}{n + 1}\\ 0 & \textrm{otherwise}\end{cases}$
\STATE ${\bf p}_c^{t+1} \gets \left( 1 - c_{c} \right) {\bf p}_c^t + h_\sigma \sqrt{c_c (2 - c_c) \mu_{eff}} {\bf y}$
  %\COMMENT{Covariance evolution path}
\STATE $c_{1a} \gets c_1 \left(1 - (1 - h_\sigma) c_c (2 - c_c) \right)$
\STATE $\boldsymbol{S} \gets \sum_{i=1}^\lambda \left(\boldsymbol{\theta}_i - {\bf W}^t \phi(\boldsymbol{s}_i)\right) \frac{{\bf D}_{ii}}{{\sigma^t}^2} \left(\boldsymbol{\theta}_i - {\bf W}^t \phi({\bf s}_i)\right)^T$
\STATE \begin{eqnarray*}
{\bf \Sigma}^{t+1} &\gets& (1 - c_{1a} - c_\mu) {\bf \Sigma}^t\\
&& + c_1 {\bf p}_c^{t+1} {{\bf p}_c^{t+1}}^T + c_\mu \boldsymbol{S}
\end{eqnarray*}
\STATE $\sigma^{t+1} \gets \sigma^t \exp \left( \frac{c_\sigma}{d_\sigma} \left( \frac{||{\bf p}_\sigma^{t+1}||}{\mathbb{E}||\mathcal{N}(0, {\bf I})||} - 1 \right) \right)$
\STATE $t \gets t + 1$
\ENDWHILE
\end{algorithmic}
\end{algorithm}

\begin{figure}[tb]
\vskip 0.2in
\begin{center}
\centerline{\includegraphics[width=0.95\columnwidth]{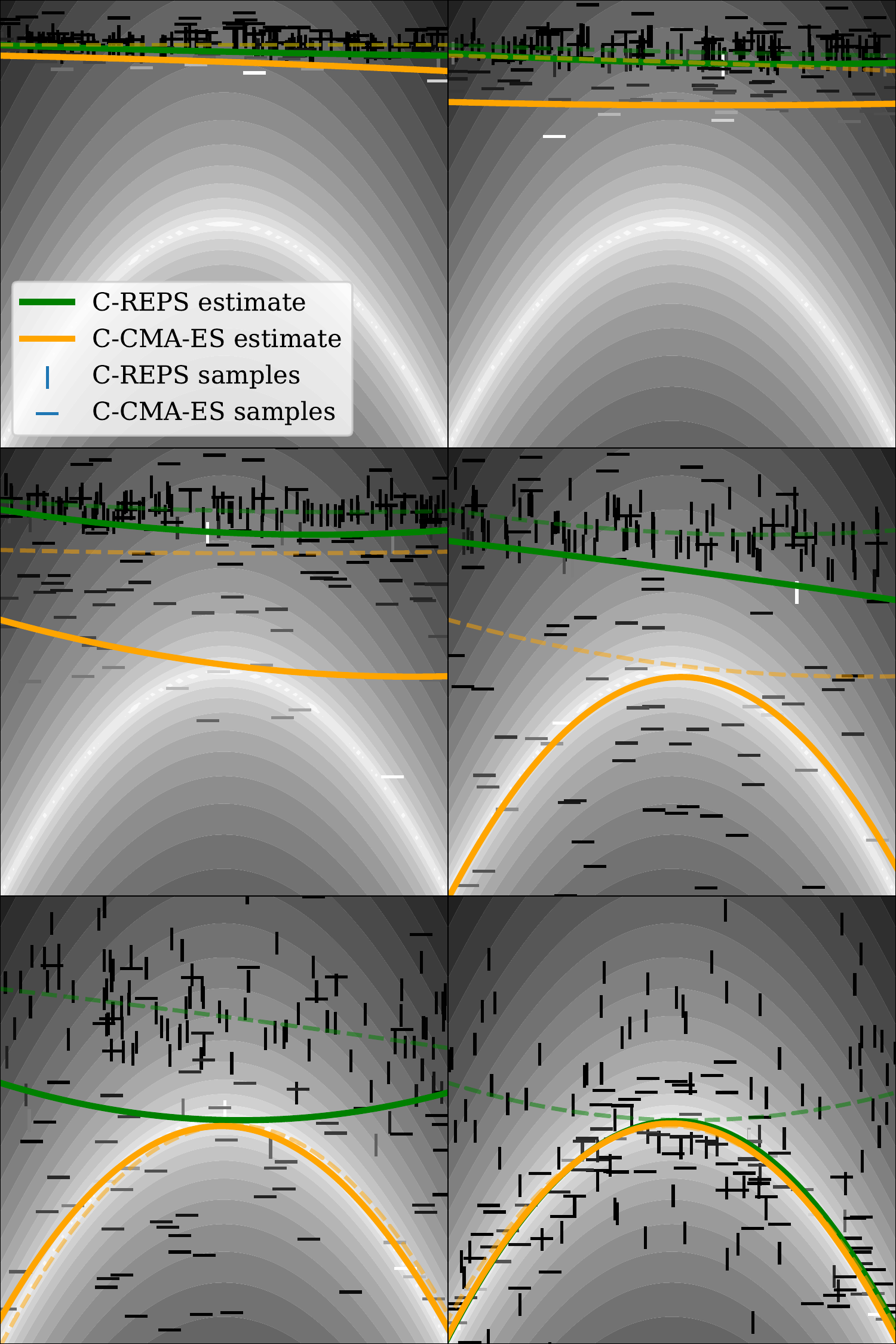}}
\caption{Comparison of C-REPS and C-CMA-ES in a simple contextual function
optimization problem in six generations. Values of the contextual objective
are indicated by background color. The optimum is a quadratic function in the
valley. The x-axis represents the context $s$ and the y-axis the parameter $x$.
In each generation, 100 samples are used to move the search distribution
indicated by the mean function from the dashed line to the solid line.}
\label{fig:illustration}
\end{center}
\vskip -0.2in
\end{figure}

C-CMA-ES is shown in Algorithm \ref{alg:CCMAES} and its default hyperparameters
in Algorithm \ref{alg:params}. We list the algorithm here, because
the original publication does not give a complete and correct listing
of the algorithm. For a more detailed description of the algorithm with more
explanations, however, please refer to \citet{Abdolmaleki2017}.
Figure \ref{fig:illustration} illustrates
how C-CMA-ES compares to C-REPS in a very simple contextual optimization
problem. The initial variance of the search distribution was set intentionally
low to demonstrate that C-CMA-ES quickly adapts its step size, whereas C-REPS
restricts the maximum Kullback-Leibler divergence between successive search
distributions which results in slow adaptation of the step size.

\begin{algorithm}[tb!]
\caption{Hyperparameters of C-CMA-ES}
\label{alg:params}
\begin{algorithmic}[1]
\REQUIRE number of samples per update $\lambda$,
sample weights $\boldsymbol{D}$,
number of parameters $n$,
number of context variables $n_s$
\STATE $\mu_{eff} \gets \frac{1}{\sum_{i=1}^\lambda {\bf D}_{ii}^2}$
\STATE $c_1 \gets 2 / \left((n + n_s + 1.3)^2 + \mu_{eff}\right)$
\STATE $c_\mu \gets \min \left( 1 - c_1, \frac{2 \left(\mu_{eff} - 2 + \frac{1}{\mu_{eff}}\right)}{\left(n + n_s + 2\right)^2 + \mu_{eff}} \right)$
\STATE $c_c \gets \frac{4 + \frac{\mu_{eff}}{n + n_s}}{4 + n + n_s + 2 \frac{\mu_{eff}}{n + n_s}}$
\STATE $c_\sigma \gets \frac{\mu_{eff} + 2}{n + n_s + \mu_{eff} + 5}$
\STATE $d_\sigma \gets 1 + 2 \max \left( 0, \sqrt{\frac{\left(\mu_{eff} - 1\right)}{(n + n_s + 1)}} - 1 \right) + c_\sigma + \log (n + n_s + 1)$
\STATE $\mathbb{E}||\mathcal{N}(0, {\bf I})|| \gets \sqrt{n} \left( 1 - \frac{1}{4 n} + \frac{1}{21 n^2} \right)$
\end{algorithmic}
\end{algorithm}

\subsection{Active C-CMA-ES}

Active CMA-ES \cite{Jastrebski2006} is an extension of CMA-ES. The covariance
update is modified to take into account the worst samples of a generation.
Similar to the rank-$\mu$ update of the covariance with the best samples of
a generation, we compute a matrix
$$
\boldsymbol{S}^- \gets \frac{1}{{\sigma^t}^2}\sum_{i=1}^\lambda \left(\boldsymbol{\theta}_j - {\bf W}^t \phi(\boldsymbol{s}_j)\right)
    \cdot {\bf D}_{ii}
 \cdot \left(\boldsymbol{\theta}_j - {\bf W}^t \phi({\bf s}_j)\right)^T,
$$
where $j = {1 + \lambda - i}$. $\boldsymbol{S}^-$ will be subtracted from
the covariance. Line 22 of Algorithm \ref{alg:CCMAES} is replaced by
\begin{eqnarray*}
{\bf \Sigma}^{t+1} &\gets& (1 - c_{1a} - c_\mu \textcolor{red}{- \frac{1}{2} c_{\mu^-}}) {\bf \Sigma}^t\\
&& + c_1 {\bf p}_c^{t+1} {{\bf p}_c^{t+1}}^T + \left(c_\mu \textcolor{red}{+ \frac{1}{2} c_{\mu^-}}\right) \boldsymbol{S} \textcolor{red}{- c_{\mu^-} \boldsymbol{S}^-},
\end{eqnarray*}
where $c_{\mu^-} = \frac{(1 - c_\mu) \mu_{eff}}{4 \left((n + n_s + 2)^{1.5} + 2 \mu_{eff}\right)}$.

\subsection{C-ACM-ES}

Another extension to CMA-ES is ACM-ES \cite{Loshchilov2010}. It uses a
ranking SVM as surrogate model to improve sample-efficiency. This integrates
well because CMA-ES only takes into account the ranking of samples in a
generation. It does not consider actual returns.
Assuming all samples are ordered by their rank, a ranking SVM minimizes the
objective
\begin{eqnarray*}
&&\frac{1}{2} ||w||_2^2 + \sum_{i=1}^N C_i \xi_i\\
&\textrm{subject to}& w^T \left(\phi(x_i) - \phi(x_{i+1})\right) \geq 1 - \xi_i \wedge
\xi_i \geq 0,\\
&&\forall i=1, \ldots, N-1.
\end{eqnarray*}
This can be solved by a form of sequential minimal optimization \citep{Platt1998}
and we can use the kernel trick. In this paper, we will use an RBF kernel
$$k(x, x') = \exp\left( -\frac{(x - x')^2}{2 \sigma^2} \right),$$
with $\sigma$ set to the average distance between training samples.
The cost of an error depends on the ranks of corresponding samples and is
$C_i = 10^6 (N-i)^{c_{pow}}$, where usually $c_{pow}=2$.

Instead of ordering samples by their returns, we will order them by samples of
their advantage values (returns with subtracted context-dependent baseline, see
Algorithm \ref{alg:CCMAES}, line 11). We found this to be crucial in
preliminary experiments.
%\todo{we could take a closer look at how ranking values and baselines interfere}

\citet{Loshchilov2010} use the surrogate only conservatively
to pre-screen a larger set of samples from which more promising samples are
selected with a higher probability for evaluation.
This is more difficult in a contextual setting because we typically have no
control over the contexts in which we can evaluate samples. Once we know in
which context we can evaluate the next sample, we could sample several
parameters vectors $\boldsymbol{\theta}_i$ and select the most promising
samples with higher probability for evaluation. In experiments that we
conducted, this often caused preliminary convergence.
Instead, we will exploit the surrogate model directly, that is, we will not use
it for pre-screening but we will use predicted ranking values directly in the
update step.

Another key idea of ACM-ES is to normalize samples
$$\boldsymbol{\theta}' \gets {\boldsymbol{\Sigma}^t}^{-\frac{1}{2}} \left(\boldsymbol{\theta} - \boldsymbol{\mu}^t\right)$$
based on the covariance and mean of the search distribution to learn the
surrogate model.
We adopt the idea of using a ranking SVM as a surrogate model with normalized
samples for C-ACM-ES. Instead of only the parameters $\boldsymbol{\theta}$,
the surrogate model will also take into account context $\boldsymbol{s}$. The
normalization is a little bit more complicated:
$$\boldsymbol{\theta}' \gets {\boldsymbol{\Sigma}^t}^{-\frac{1}{2}} \left(\boldsymbol{\theta} - {\boldsymbol{W}^t}^T \phi(\boldsymbol{s})\right),$$
and the contexts of the training set will be normalized to have a mean of zero
and a standard deviation of one. In this paper, we assume that there is no
correlation between context variables, which is not correct in general.

The search distribution is updated after $\lambda$ samples from the objective function.
We store the last $40 + \lfloor 4 d^{1.7} \rfloor$ samples to train the local
surrogate model, where $d$ is the number of parameters to be optimized.
For each update of the search distribution, in addition to the $\lambda$ samples
that we evaluated on the real objective function, we will draw
$\lambda' - \lambda$ samples from the previous search distribution for random
contexts that we observed in the training set and predict their ranking values
to compute the update of the search distribution with these $\lambda'$ samples.
Additional hyperparameters will be described and evaluated in Section
\ref{sec:hyperparameters}.
%\todo{pre-screening did not work, explain the algorithm in detail}

Using a surrogate model does not decrease the computational complexity
in comparison to C-CMA-ES. Our expectation is that it increases
sample-efficiency and, hence, the suitability for expensive objective
functions.

\section{Evaluation}

We will evaluate the described extensions in contextual black-box optimization
and two deterministic, simulated robotic problems.

\subsection{Contextual Black-box Optimization}

Another idea that can be transferred from black-box optimization to the
contextual setting is a set of standard benchmark functions.
The analysis in this section is very similar to the one of \citet{Abdolmaleki2017}.
We use some additional objective functions. We take standard objective functions
and make them contextual by defining
$f_{\boldsymbol{s}}(\boldsymbol{\theta}) = f(\boldsymbol{\theta} + \boldsymbol{G} \phi(\boldsymbol{s}))$,
where $\boldsymbol{G}$ is a matrix with components samples iid from a
standard normal distribution.
If not stated otherwise, $n_s=1$,
$\phi(\boldsymbol{s}) = \boldsymbol{s}$, and the components of $\boldsymbol{s}$
are sampled from the interval $\left[1, 2\right)$.

To make results comparable to the one of \citet{Abdolmaleki2017},
we use the same definition of
% x.dot(x)
$f_{Sphere}(x) = \sum_{i=1}^d x_i^2$
and
% np.sum(100 * (x[:-1] ** 2 - x[1:]) ** 2 + (x[:-1] - 1) ** 2)
$f_{Rosenbrock}(x) = \sum_{i=1}^{d-1} 100 (x_i^2 - x_{i+1})^2 + (x_i - 1)^2$.
In addition, we use
% f = (-20.0 * np.exp(-0.2 * np.sqrt(np.mean(x ** 2))) - np.exp(np.mean(np.cos(2.0 * np.pi * x))) + 20 + np.exp(1.0))
\begin{eqnarray*}
f_{Ackley}(x) &=& -20 \exp \left(-0.2 \sqrt{\frac{1}{d}\sum_{i=1}^d x_i^2} \right) + 20\\
&& - \exp \left(\frac{1}{d}\sum_{i=1}^d \cos(2 \pi x_i) \right) + \exp(1)
\end{eqnarray*}
and the functions \textit{ellipsoidal}, \textit{discus}, and \textit{different powers}
from the COCO platform, a benchmark platform for black-box
optimization \cite{Hansen2016}.
The sphere objective checks the optimal convergence rate of an
algorithm, the ellipsoidal function has a high conditioning that
requires the algorithm to estimate the individual learning rates per
dimension but is symmetric and separable, the Rosenbrock function checks
whether the optimizer is able to change its direction multiple times,
the discus function also has a high conditioning, the different powers
function has no self-similarity, and the Ackley function is a multimodal
function with many local minima.
We do not use any other multimodal function because we
do not have any contextual black-box optimization
algorithm that works well for these kind of problems.
They are particularly challenging for contextual
optimizers.
% because we could end up in different local minima
%in different contexts from which it is very difficult to escape
%and the solution between those contexts will not even be in a
%local minimum.

Initial parameters are sampled from $\mathcal{N}(\boldsymbol{0}, \sigma_0^2\boldsymbol{I})$
with $\sigma_0=1$ in most cases. Initial parameters of the Ackley function are sampled
with $\sigma_0=14.5$ and in general the function has bounds at $\left[-32.5, 32.5\right]$. The number of dimensions of
the parameter vector $\boldsymbol{\theta}$ is 20.
$\lambda$ is set close to the smallest possible value that generates a
stable learning progress. This is also the case in all following experiments.
Unless otherwise stated, we use $\lambda=50$ samples of the objective function before
we make an update of the search distribution. The number of updates corresponds
to the number of iterations or generations in the following analysis.
Instead of minimizing the $f_{\boldsymbol{s}}$, we will maximize
$-f_{\boldsymbol{s}}$. Listed function values are averaged over one generation,
that is, multiple contexts will be considered but not always the same contexts.

\subsubsection{Hyperparameters of C-ACM-ES}
\label{sec:hyperparameters}

First, we try to find a good configuration of C-ACM-ES.
There are several hyperparameters:
we have to define after how many samples the surrogate
model is accurate enough to be used ($n_{start}$),
the number of samples $\lambda'$ evaluated by the
surrogate model, $c_{pow}$ of the ranking SVM objective,
and the number of iterations
$n_{iter}$ that will be used to optimize the ranking SVM
per training sample.
While one parameter has been investigated,
the others were kept to the values $\lambda' = 3\lambda$,
$n_{start}=100$, % actually n_{start} = 3000 for c_{pow}
$c_{pow}=1$, $n_{iter}=1000$.
We found that $n_{start}$ is particularly important for optimizing
$f_{Ackley}$ and $c_{pow}$, $n_{iter}$, and $\lambda'$ are important for
$f_{Rosenbrock}$.

The most important results of the performed experiments are shown in
Table \ref{tab:test_params}. Setting $\lambda'= 2 \lambda$ gives the
best result for the Rosenbrock function. However, that
$\lambda' = 3 \lambda$ is a better compromise between exploitation of
the model and a conservative handling. In fact, there are
some objectives, where we can much better exploit the model. In the
following experiments, we will use the configurations $\lambda'= 3 \lambda$
and $\lambda' = 10 \lambda$. Larger values for $n_{iter}$
improve the result, which is not surprising. This is especially the case on
a complex function like the Rosenbrock function. As a compromise between
computational overhead and sample-efficiency, we select $n_{iter}=1000$
which seems to work reasonably well. $n_{iter}=3000$ also does not seem
to be a bad choice as it significantly improves the result on the Rosenbrock
function and only increases computational cost by a factor of three.
On the Ackley function, it is important to bring the search distribution
in a good state in which we can learn an accurate surrogate model before
we start exploiting the surrogate. $n_{start} = 3000$ seems to be the best
parameter here. In addition, we will use an aggressive version in the
following experiments and set $n_{start} = 100$. The effect of $c_{pow}$
is quite interesting. Although in the original implementation for ACM-ES
\cite{Loshchilov2010} the default value is 2, this seems to have a
catastrophic effect on the Rosenbrock function. For all other functions,
differences are negligible. During all conducted experiments,
we found the estimate of the context-dependent baseline and
the surrogate model to be very critical for C-ACM-ES to function.

We also tried kernel ridge regression with the same kernel as
for the ranking SVM to learn a surrogate that estimates expected
return. However, the results were not promising as suggested already
by \citet{Loshchilov2010}.

\begin{table}[t]
\caption{Comparison of hyperparameters, average of 20 runs.}
\label{tab:test_params}
\vskip 0.1in
\begin{center}
\begin{small}
\begin{sc}
\begin{tabular}{ll}
\toprule
Hyperparam. & $\frac{1}{|S|}\sum_{\boldsymbol{s} \in S} f_{\boldsymbol{s}}(x)$ \\
\midrule
\multicolumn{2}{c}{Rosenbrock ($n_s=1$), after generation 850}\\
\midrule
$\lambda' = 2 \lambda$ & \boldmath$-7.817 \cdot 10^{-10}$\\
$\lambda' = 3 \lambda$ & $-1.485 \cdot 10^{-9}$\\
$\lambda' = 5 \lambda$ & $-4.089 \cdot 10^{-3}$\\
$\lambda' = 10 \lambda$ & $-1.445 \cdot 10^{15}$\\
\midrule
$n_{iter} = 300$ & $-1.679 \cdot 10^{-3}$\\
$n_{iter} = 1000$ & $-1.485 \cdot 10^{-9}$\\
$n_{iter} = 300$ & $-4.607 \cdot 10^{-13}$\\
$n_{iter} = 10000$ & \boldmath$-2.396 \cdot 10^{-14}$\\
\midrule
$c_{pow} = 1$ & \boldmath$-3.656 \cdot 10^{-9}$\\
$c_{pow} = 2$ & $-1.977 \cdot 10^{41}$\\
%$c_{pow} = 1$ & Diff. Powers & 600 & \boldmath$-2.646 \cdot 10^{-14}$\\
%$c_{pow} = 2$ & Diff. Powers & 600 & $-5.810 \cdot 10^{-14}$\\
%$c_{pow} = 1$ & Ackley & 1100 & \boldmath$-3.995 \cdot 10^{-9}$\\
%$c_{pow} = 2$ & Ackley & 1100 & $-4.954 \cdot 10^{-9}$\\
%$c_{pow} = 1$ & Ellipsoidal & 800 & \boldmath$-1.039 \cdot 10^{-10}$\\
%$c_{pow} = 2$ & Ellipsoidal & 800 & $-1.107 \cdot 10^{-10}$\\
\midrule
\multicolumn{2}{c}{Ackley ($n_s=1$), after generation 1100}\\
\midrule
$n_{start} = 100$& $-1.411 \cdot 10^1$\\
$n_{start} = 300$ & $-1.085 \cdot 10^1$\\
$n_{start} = 1000$ & $-1.086 \cdot 10^0$\\
$n_{start} = 3000$ & \boldmath$-3.995 \cdot 10^{-9}$\\
$n_{start} = 10000$ & $-1.155 \cdot 10^{-8}$\\
\bottomrule
\end{tabular}
\end{sc}
\end{small}
\end{center}
\vskip -0.1in
\end{table}

\subsubsection{Comparison of C-CMA-ES Extensions}

\begin{figure}[tb!]
\vskip 0.2in
\begin{center}
\centerline{\includegraphics[width=\columnwidth]{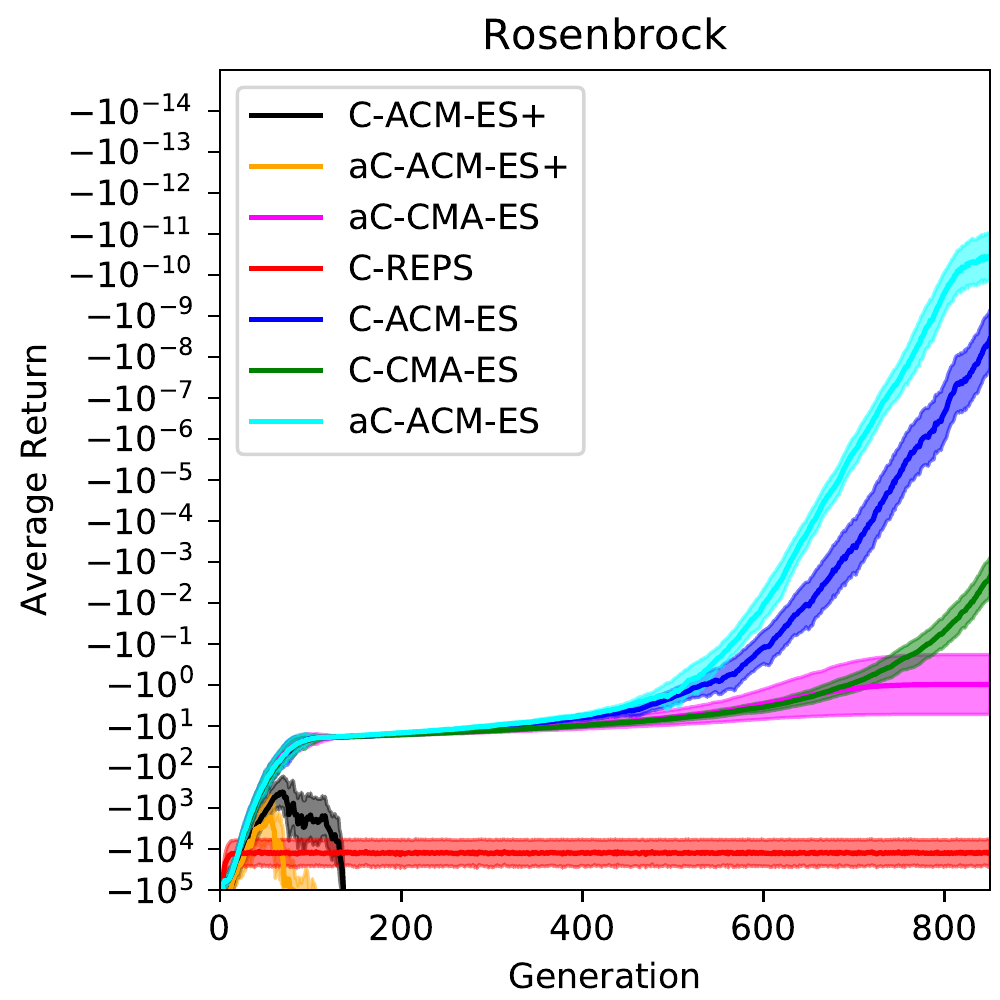}}
\caption{Learning curves for the Rosenbrock function of several contextual
policy search methods. Mean and standard deviation of 20 experiments are
displayed.}
\label{fig:rosenbrock}
\end{center}
\vskip -0.2in
\end{figure}

We did an extensive evaluation of several combinations of extensions of
C-CMA-ES. Results are displayed in Table \ref{tab:test}.
NaN indicates divergence.
We use C-REPS with the hyperparameter $\epsilon=1$ and C-CMA-ES as baselines.
Note that it is usually better to set $\epsilon$ for C-REPS as large as
possible to avoid premature convergence, however, the algorithm becomes
numerically instable if $\epsilon$ is too large. $\epsilon=1$ is a good
trade-off. In more real-world scenarios, setting $\epsilon$ to such a
large value (or setting the initial step size of CMA-ES to a large value)
could result in dangerous exploration.
The term aC-CMA-ES refers to active C-CMA-ES, C-ACM-ES uses the
surrogate model, and aC-ACM-ES is its active counterpart.
``+'' indicates that the surrogate model is exploited
aggressively, that is, we set $\lambda' = 10 \lambda$ and $n_{start} = 100$,
otherwise $\lambda' = 3 \lambda$ and $n_{start} = 3000$. Exemplary
learning curves are displayed in Figure \ref{fig:rosenbrock} for the
Rosenbrock function.

\begin{figure}[tb!]
\vskip 0.2in
\begin{center}
\centerline{\includegraphics[width=\columnwidth]{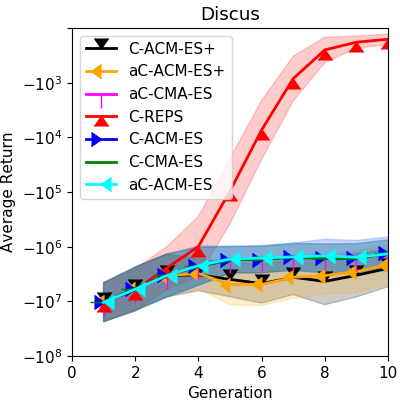}}
\caption{Learning curves for the Discus function for the first few generations.
Mean and standard deviation of 20 experiments are displayed.}
\label{fig:discus}
\end{center}
\vskip -0.2in
\end{figure}

Variants of C-ACM-ES usually outperform vanilla C-CMA-ES.
Although the surrogate model focuses on ordering the samples with the highest
rank more correctly and aC-CMA-ES is often not better than C-CMA-ES,
aC-ACM-ES performs best in most cases. If this is not the case, C-ACM-ES
performs best.
The sphere function with two context variables seems to be different. Here,
it is important to exploit the surrogate model as early and aggressively as
possible to have a chance against C-CMA-ES.

\begin{table}[tb!]
\caption{Comparison of CPS methods, average of 20 runs. Best results are
underlined.}
\label{tab:test}
\vskip 0.1in
\begin{center}
\begin{small}
\begin{sc}
\begin{tabular}{rr}
\toprule
Method & $\frac{1}{|S|}\sum_{\boldsymbol{s} \in S} f_{\boldsymbol{s}}(x)$ \\
\midrule
\multicolumn{2}{c}{Sphere ($n_s=2$), after generation 200}\\
\midrule
C-REPS & $-4.509 \cdot 10^{+01}$\\
C-CMA-ES & $-1.815 \cdot 10^{-05}$\\
aC-CMA-ES & $-1.348 \cdot 10^{-05}$\\
\underline{C-ACM-ES+} & \underline{$-1.294 \cdot 10^{-08}$}\\
aC-ACM-ES+ & $-1.506 \cdot 10^{-01}$\\
C-ACM-ES & $-6.257 \cdot 10^{-04}$\\
aC-ACM-ES & $-2.309 \cdot 10^{-04}$\\
\midrule
\multicolumn{2}{c}{Rosenbrock ($n_s=1$), after generation 850}\\
\midrule
C-REPS & $-1.255 \cdot 10^{+04}$\\
C-CMA-ES & $-2.328 \cdot 10^{-03}$\\
aC-CMA-ES & $-9.736 \cdot 10^{-01}$\\
C-ACM-ES+ & $-1.445 \cdot 10^{+15}$\\
aC-ACM-ES+ & $-3.227 \cdot 10^{+19}$\\
C-ACM-ES & $-3.656 \cdot 10^{-09}$\\
\underline{aC-ACM-ES} & \underline{$-3.899 \cdot 10^{-11}$}\\
\midrule
\multicolumn{2}{c}{Ackley ($n_s=1$), after generation 1100}\\
\midrule
C-REPS & $-1.947 \cdot 10^{+01}$\\
C-CMA-ES & $-8.762 \cdot 10^{-07}$\\
aC-CMA-ES & $-8.773 \cdot 10^{-07}$\\
C-ACM-ES+ & NaN\\
aC-ACM-ES+ & NaN\\
\underline{C-ACM-ES} & \underline{$-3.995 \cdot 10^{-09}$}\\
aC-ACM-ES & $-1.813 \cdot 10^{-08}$\\
\midrule
\multicolumn{2}{c}{Ellipsoidal ($n_s=1$), after generation 800}\\
\midrule
C-REPS & $-2.944 \cdot 10^{+05}$\\
C-CMA-ES & $-2.337 \cdot 10^{+02}$\\
aC-CMA-ES & $-1.524 \cdot 10^{+02}$\\
C-ACM-ES+ & $-1.300 \cdot 10^{+16}$\\
aC-ACM-ES+ & $-2.407 \cdot 10^{+18}$\\
C-ACM-ES & $-1.039 \cdot 10^{-10}$\\
\underline{aC-ACM-ES} & \underline{$-2.388 \cdot 10^{-11}$}\\
\midrule
\multicolumn{2}{c}{Diff. Powers ($n_s=1$), after generation 600}\\
\midrule
C-REPS & $-9.088 \cdot 10^{+02}$\\
C-CMA-ES & $-1.562 \cdot 10^{-07}$\\
aC-CMA-ES & $-3.038 \cdot 10^{-07}$\\
C-ACM-ES+ & $-7.111 \cdot 10^{+74}$\\
aC-ACM-ES+ & $-8.717 \cdot 10^{+82}$\\
C-ACM-ES & $-2.464 \cdot 10^{-14}$\\
\underline{aC-ACM-ES} & \underline{$-1.284 \cdot 10^{-14}$}\\
\midrule
\multicolumn{2}{c}{Discus ($n_s=1$), after generation 850}\\
\midrule
C-REPS & $-1.288 \cdot 10^{+02}$\\
C-CMA-ES & $-2.995 \cdot 10^{-10}$\\
aC-CMA-ES & $-3.838 \cdot 10^{-10}$\\
C-ACM-ES+ & $-8.297 \cdot 10^{+27}$\\
aC-ACM-ES+ & $-1.250 \cdot 10^{+24}$\\
\underline{C-ACM-ES} & \underline{$-8.877 \cdot 10^{-12}$}\\
aC-ACM-ES & $-1.684 \cdot 10^{-11}$\\
\bottomrule
\end{tabular}
\end{sc}
\end{small}
\end{center}
\vskip -0.1in
\end{table}

An interesting result, however, is that C-REPS is often much faster in the
early phase. See, for example, Figure \ref{fig:discus}. In the first
10 generations, which amounts to 500 episodes, C-REPS outperforms all
variants of C-CMA-ES by orders of magnitude. Unfortunately this is the
phase of the optimization that is usually interesting for learning in
the real world. We can also see that C-REPS converges already and does not
learn anything for the next 840 generations. Variants of C-CMA-ES will
continue making progress until the last episode.
Because the surrogate model is only useful when we have a good estimate
of the covariance matrix and a substantial amount of samples from the
objective function, we can only use it in later stages of the optimization to
improve the learning progress.

\subsection{2D Viapoint Problem}

\begin{figure}[tb]
\vskip 0.2in
\begin{center}
\centerline{\includegraphics[width=0.95\columnwidth]{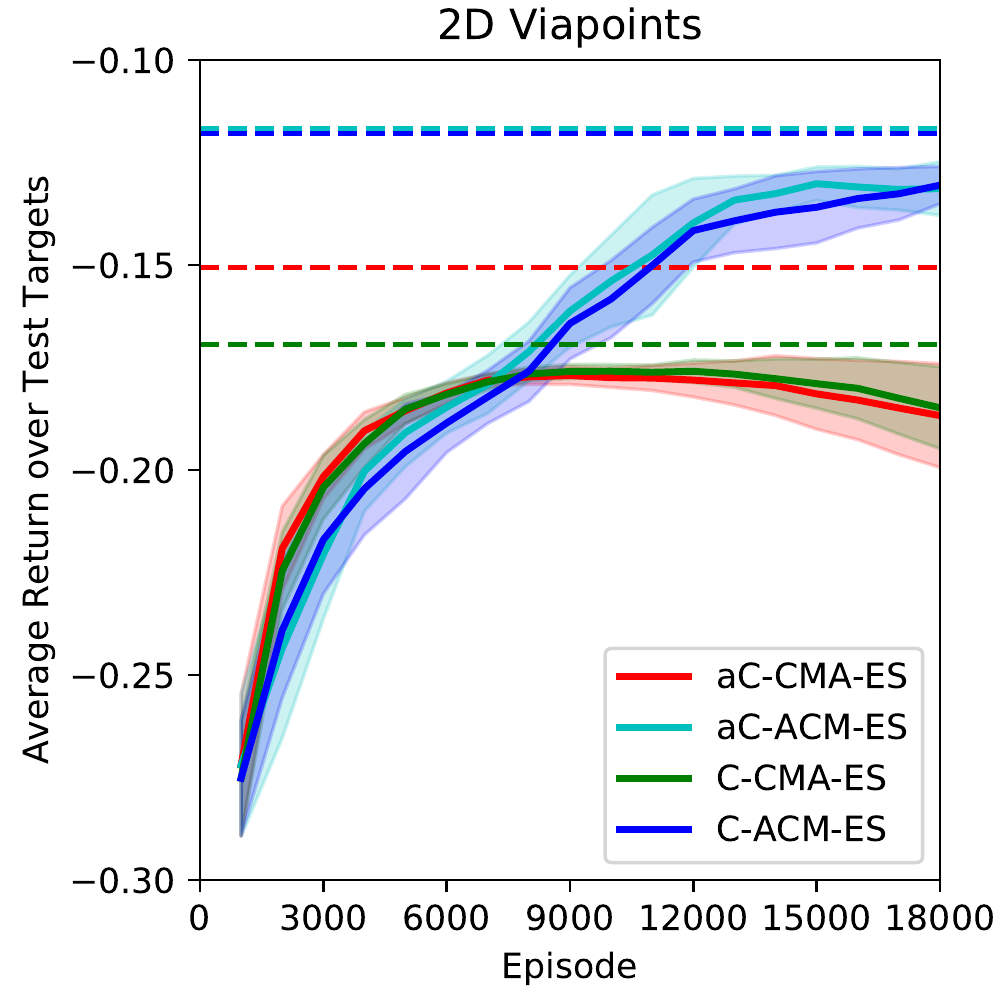}}
\caption{Learning curves for the viapoint problem, averaged over 20 experiments.
Dashed lines indicate the maximum return over all experiments and generations.}
\label{fig:viapoint}
\end{center}
\vskip -0.2in
\end{figure}

We will use a 2D viapoint problem. A dynamical movement primitive
\cite{Ijspeert2013} with
$\boldsymbol{x}_0 = (0, 0)^T$, $\boldsymbol{g} = (1, 1)^T$, $\tau=1.0$,
and 10 parameters per dimension will be used as trajectory representation.
The reward is defined for each step $t = 0, \ldots, T-1$ as
$$r_t = -0.001||\boldsymbol{v}||_t,$$
and for the last step as the difference to each viapoint
$$r_T = -\sum_{\boldsymbol{p}_{via,t}} ||\boldsymbol{p}_{via,t} - \boldsymbol{p}_t||.$$
Viapoints are defined as a tuple of time and positions
$\{ \left(0.2, (0.2, 0.5)^T \right), \left(0.5, \boldsymbol{s} \right) \}$,
where $\boldsymbol{s} \in \left[ 0.3, 0.7 \right] \times \left[ 0.3, 0.7 \right]$.

We did not use C-REPS as a baseline because it is not robust against
the choice of its hyperparameter $\epsilon$ \cite{Fabisch2015}.
In the experiments, we use $\lambda=100$, $n_{start} = 1000$, and a quadratic
baseline.
Learning curves are displayed in Figure \ref{fig:viapoint}. The performance
is evaluated on a grid of 25 test contexts, where
$\boldsymbol{s}_1, \boldsymbol{s}_2 \in (0.3, 0.4, 0.5, 0.6, 0.7)$.
Active C-CMA-ES does not make any difference.
The convergence behavior of (a)C-ACM-ES is much better, however, the advantage
only occurs after about 8,000 episodes, which is way too late for such a simple
problem from the robotics perspective.

\subsection{Ball Throwing}

\begin{figure}[tb]
\vskip 0.2in
\begin{center}
\centerline{\includegraphics[width=0.95\columnwidth]{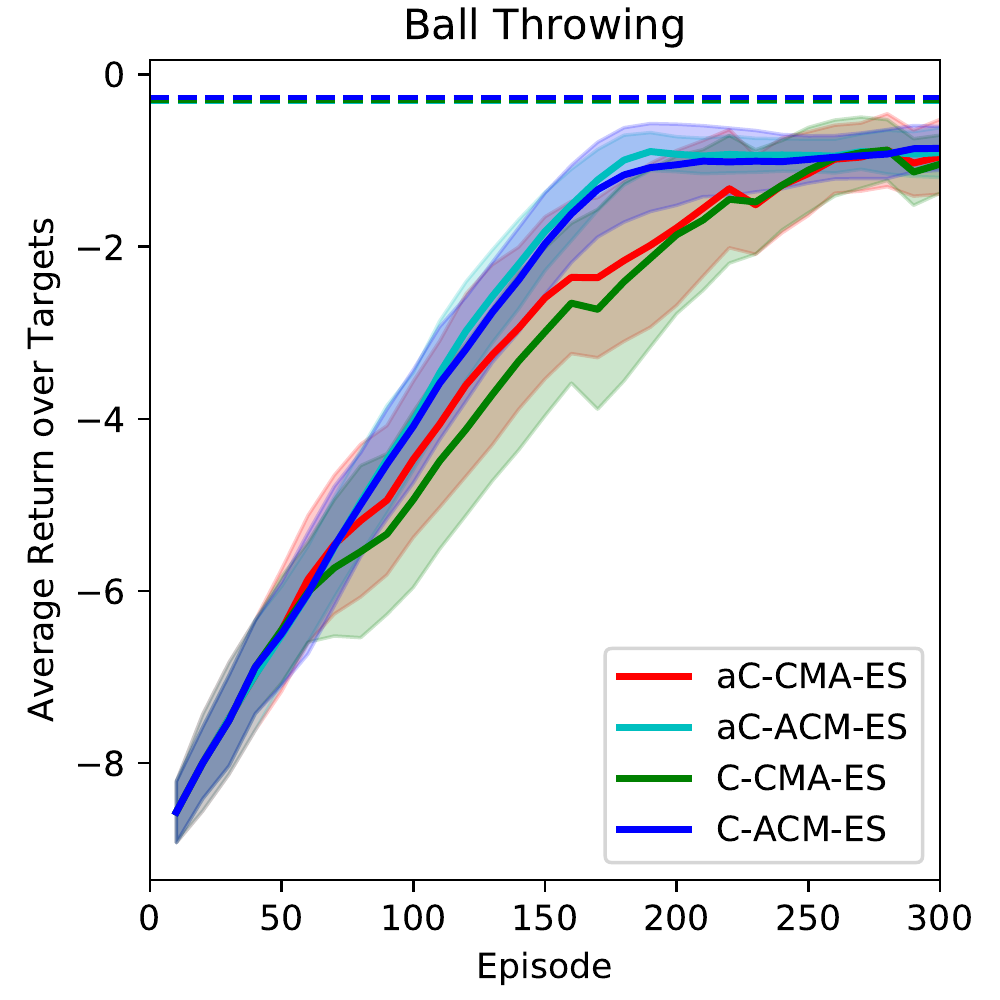}}
\caption{Learning curves for the ball throwing problem, averaged over 20 experiments.
Dashed lines indicate the maximum return over all experiments and generations.}
\label{fig:ball_throwing}
\end{center}
\vskip -0.2in
\end{figure}

An example of a problem where it is much easier to define the
reward function than the solution, is throwing a ball at a target.
A reward is only given at the end of an episode, hence, we can directly
define the return
$$R(\boldsymbol{s}, \boldsymbol{\theta}) = -\max_{i,t} |\dot{\boldsymbol{q}}_{i,t}| - ||\boldsymbol{s} - \boldsymbol{p}||_2,$$
where $\max_{i,t} |\dot{\boldsymbol{q}}_{i,t}|$ is the maximum velocity in a
joint during execution of a robot's throwing motion and $\boldsymbol{p}$ is
the point where the ball hits the ground.
We define a problem which is very similar to the ball-throwing problem
of \citet{Fabisch2015}: a dynamical movement primitive represents a
throwing movement of a 6 DOF robot with 10 weights per dimension. An initial
movement hits the target $(10, 5)^T$. We try to generalize only over
three arbitrarily selected contexts / target positions:
$\{(2, 3)^T, (1.5, 2.2)^T, (1, 2)^T\}$, set $n_{start}=40$, and $\lambda=10$.

Results are shown in Figure \ref{fig:ball_throwing}. The learning curve is
very steep because the initial trajectory has to be adapted a lot to hit
the new targets. We can set $n_{start}$ and $\lambda$ to very low values
because we only want to generalize over a set of three discrete contexts.
In this case, using a surrogate model already gives a slight advantage so
early in the learning process. The difference between active and standard
covariance updates is not significant.

\section{Conclusion and Outlook}

% Summary
We demonstrated that the extensions active C-CMA-ES and C-ACM-ES can
be combined and yield impressive results on contextual function optimization
problems in comparison to C-CMA-ES.
We have shown, however, that these results are actually not directly
transferable to the domain of robotics.
We would like to learn successful upper-level policies
in 100--1000 episodes at maximum. The presented extensions, however,
start to be better than standard C-CMA-ES just after the range of interest.
They exhibit much better convergence behavior though.

% Outlook
A drawback of many contextual optimization algorithms (for example, C-REPS and
C-CMA-ES) is that they learn linear upper-level policies. The choice of $\phi$
defines the functions that can be represented. BO-CPS does not have
this drawback because the policy is represented implicitly as an optimization
problem over a global surrogate model similar to how policies are defined in
value iteration reinforcement learning or Q-learning. This is expensive
to compute though. More complex upper-level policies in C-CMA-ES would
be an option to mitigate this restriction. We cannot say yet if more
complex models are stable enough to be learned in practice.

\begin{acks}
This work received funding from the European Union's Horizon 2020
research and innovation program under grant agreement No 723853.
\end{acks}

%%% -*-BibTeX-*-
%%% Do NOT edit. File created by BibTeX with style
%%% ACM-Reference-Format-Journals [18-Jan-2012].

\end{document}